\pdfoutput=1

\documentclass[11pt]{article}

\usepackage{hyperref}
\hypersetup{
    colorlinks,
    linkcolor={red!50!black},
    citecolor={green!50!black},
    urlcolor={blue!80!black}
}

\usepackage[preprint]{acl}

\usepackage{times}
\usepackage{latexsym}

\usepackage[T1]{fontenc}

\usepackage[utf8]{inputenc}

\usepackage{microtype}

\usepackage{inconsolata}

\usepackage{graphicx}

\usepackage{listings}
\usepackage{fancyvrb}
\usepackage{fvextra}

\usepackage{amssymb}
\usepackage{amsmath}
\usepackage{bm}
\usepackage{booktabs}
\usepackage{breakcites}
\usepackage{caption}
\usepackage{subcaption}
\usepackage[capitalize]{cleveref}
\usepackage[inline, shortlabels]{enumitem}
\usepackage{placeins}
\usepackage{mathtools}
\usepackage{tabularx}
\usepackage{siunitx}
\usepackage{verbatim}
\usepackage{pifont}
\usepackage{wrapfig}
\usepackage{multicol}
\usepackage[inkscapelatex=false]{svg}
\usepackage[multiple]{footmisc}
\usepackage[color=yellow]{todonotes}
\usepackage{xcolor}

\newcommand{\lmh}{h_{\text{chat}}}  
\newcommand{\gmh}{h_{\text{guard}}}  
\newcommand{\lmf}{f} 
\newcommand{\gmf}{f'} 
\newcommand{\R}{\mathbb{R}}
\usepackage{xspace} 
\newcommand{\our}{\texttt{LoRA-Guard}\xspace}

\newcommand{\tableinfo}{%
For each metric, we report the median value across 3 random seeds with the range in parentheses.
Size refers to the parameter overhead of the guard model when
run in conjunction with the corresponding chat model. }

\newlist{todolist}{itemize}{2}
\setlist[todolist]{label=$\square$}
\title{\texttt{LoRA-Guard}: Parameter-Efficient Guardrail Adaptation for Content Moderation of Large Language Models\thanks{
\textbf{Version Note: } Changes in this version v2 relative to v1: 
separate output heads for safe/unsafe classification and harm category classification (\S\ref{sec:impl}),
training on BeaverTails dataset (\S\ref{sec:beavertails}),
use of recent chat models (\S\ref{sec:chatmodels}),
comparison with recent guard models (\S\ref{sec:results}).}}
\author{
    \textbf{Hayder Elesedy}
    \quad  
    \textbf{Pedro M. Esperan\c{c}a}
    \quad 
    \textbf{Silviu Vlad Oprea}
    \quad 
    \textbf{Mete Ozay}
    \\ \\
Samsung R\&D Institute UK (SRUK), United Kingdom
\\
  \small{
    \textbf{Correspondence:} 
    \{%
    \href{mailto:p.esperanca@samsung.com}{p.esperanca}, \href{mailto:m.ozay@samsung.com}{m.ozay}%
    \}@samsung.com
  }
}

\begin{document}
\maketitle

\begin{abstract}
Guardrails have emerged as comprehensive method of content moderation for large language models (LLMs), complementing
safety alignment from fine-tuning.
However, existing model-based guardrails are too memory intensive for use on
resource-constrained computational devices such as mobile phones, an
increasing number of which are running LLM-based applications locally.
We introduce \our, a parameter-efficient guardrail adaptation method that
relies on knowledge sharing between LLMs and guardrail models. \our extracts language features from the LLMs and adapts them for the content moderation task using low-rank adapters in a dual-path design which prevents any performance degradation on the generative task.
We show that \our outperforms existing guardrail approaches while using 100-1000x fewer guardrail parameters,
enabling on-device content moderation.
\end{abstract}
\section{Introduction}
\label{section:introduction}

Large Language Models (LLMs) have become increasingly competent at language generation tasks. The standard procedure for training LLMs involves unsupervised learning of language structure from large corpora (pre-training; \citealp{achiam2023gpt4}); followed by fine-tuning on specific tasks.
For instance, conversational assistants (or chat models) are trained to respond to questions by
providing answers which are aligned with human preferences (instruction tuning; \citealp{wei2021finetuned,ouyang2022training}).

\begin{figure}[!ht]
\centering
\includegraphics[width=1.\columnwidth,trim={10px 90px 10px 10px},clip]{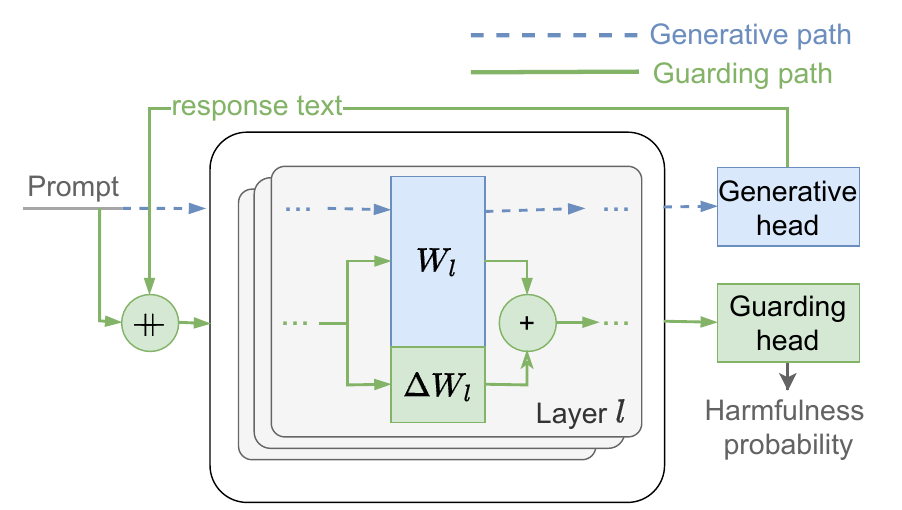}
\caption{
Overview of \our, outlined in~\cref{section:methodology}.
The generative path uses only the chat model weights ($W$) to produce a response, while the guarding path uses both the chat weights and the guard adaptors ($W$ and $\Delta W$, respectively) to produce a harmfulness score.
The system can guard the user prompt, the model response, or their concatenation ($\mathbin{+\mkern-10mu+}$). 
}
\label{fig:peft-guard-diagram}
\end{figure}
\begin{figure}[!t]
\centering
\includegraphics[width=.9\columnwidth,trim={5px 5px 5px 5px},clip]{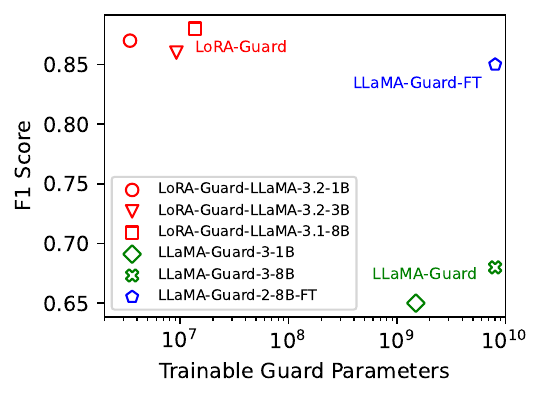}
\caption{%
Harmful content detection on BeaverTails-30k test set.
\our{} performs on-par with or better than competing guard models,
at 100-1000x reduction in guard parameters (additional to those needed to run the chat
application being monitored).
\our{} and LLaMA-Guard-2-8B-FT have been trained on BeaverTails-30k, while the 
LLaMA-Guard-3 models have not.}\label{fig:summary_results}
\end{figure}

A known failure mode of LLMs is their propensity to generate undesirable content, such as offensive language or illegal advice.
This is due to the presence of such material in their pre-training datasets, e.g., Common Crawl \citep{luccioni2021s}.
This behaviour is detrimental to safety and arises as an unintended consequence of their ability to generate helpful answers or responses which are coherent with user input \citep{wei2024jailbroken}.

To mitigate this problem, models have been optimised to not only follow instructions, but also respond in a manner that is safe and aligned with human values (safety tuning; \citealp{bai2022training,bai2022constitutional}).
However, these models are still susceptible to \textit{jailbreak} attacks, which evade the defences introduced by safety tuning via strategies such as using low-resource languages in prompts, refusal suppression, privilege escalation and distraction 
\citep{schulhoff2023ignore,dong2024attacks,shen2023anything,wei2024jailbroken}.
This has motivated the development of guardrails which monitor exchanges between chat models and users, flagging harmful entries.
Due to the failures of inbuilt safety mechanisms, guardrails form an important component of the AI safety stack in deployed systems \citep{dong2024building}.

Typically, model-based guardrails (\emph{guard models}) are distinct from the models used in the chat application being monitored~\citep{inan2023llama,ibm2024llmguard}.
However, this introduces a parameter overhead which is prohibitive in low-resource settings. 
This is inefficient: language understanding abilities of the chat models must significantly overlap those of the guard 
models if both are to effectively perform their individual tasks (response generation and content moderation, respectively).

In this paper, we propose \our, which de-duplicates these abilities via parameter sharing and parameter-efficient fine-tuning.
\our uses a low-rank adapter~(LoRA; \citealp{hu2021lora}) on the backbone transformer of the chat model to achieve a memory efficient, integrated chat and guard system.
The transformer parameters are frozen, while the LoRA parameters are trained to detect harmful content.
The LoRA parameters can be activated for guardrailing, in which case harmfullness scores are provided by a classification head, and deactivated
for chat usage, in which case the original chat model is recovered by passing the transformer outputs through the original output head.

\subsection*{Contributions}
We present \our, a parameter
efficient content moderation framework for chat applications,
allowing for guard model deployment in resource-constrained settings.
\our provides guard model systems with vast reductions in parameter overheads 
with respect to current state of the art 
(100-1000x reduction in our experiments)
while maintaining or improving content moderation performance 
(see~\cref{fig:summary_results}). 
We give performance evaluations of \our both in-distribution and 
for zero-shot generalisation on out-of-distribution data.
In~\cref{sec:low-parameter-baselines} we provide
an ablation study, finding that the LoRA adapters
are beneficial (versus just the output head or Self-Defense~\citep{phute2024selfdefense}).

\section{Background: LoRA}\label{sec:lora}
Low-Rank Adaptation~\citealp{hu2021lora}, 
or LoRA for short, is a popular method for 
parameter-efficient fine-tuning of neural networks.
LoRA is performed by freezing the weights of a pre-trained model and 
adding trainable low-rank perturbations,
replacing pre-trained weights $W \in \R^{m\times n}$ 
with $W + \frac{\alpha}{r} AB$ where 
$A\in \R^{m \times r}$, 
$B\in\R^{r\times n}$,
$r$ is the rank of the perturbations, and $\alpha$ is a scaling constant. 
During training $W$ is frozen and $A$ and $B$ are trainable parameters.
We refer to $r$, the rank of the perturbations, as the LoRA rank. 
Training the low-rank perturbations rather than the original parameters can vastly reduce the number of trainable parameters, often without
affecting performance compared to a full fine-tune~\citep{hu2021lora}.
After training, the low-rank perturbations can optionally be merged (by addition)
into the pre-trained parameters meaning 
that the fine-tuning
process incurs zero additional inference 
latency in general.
However, in this work, we 
maintain the separation between the 
LoRA perturbations 
$\Delta W = \frac{\alpha}{r} AB$ and the pre-trained
parameters
so that we may activate and deactivate the adaptation for guard and chat applications respectively.



\begin{table*}[h]
  \begin{tabular}{lllllll}
    \toprule
    Model & AUPRC$\uparrow$  & Precision$\uparrow$ & Recall$\uparrow$ & F1$\uparrow$ & FPR$\downarrow$ & Size$\downarrow$\\
     \midrule
     \our{}-\texttt{LLaMA-3.2-1B} & $.94 \;\; (.01)$ & $.88 \;\; (.04)$ & $.85 \;\; (.06)$ & $.87 \;\; (.01)$ & $.16 \;\; (.07)$ & $\num{3e06}$ \\
     \our{}-\texttt{LLaMA-3.2-3B} & $.95 \;\; (.00)$ & $.91 \;\; (.07)$ & $.82 \;\; (.15)$ & $.86 \;\; (.06)$ & $.11 \;\; (.10)$ & $\num{9e06}$ \\
     \our{}-\texttt{LLaMA-3.1-8B} & $.95 \;\; (.01)$ & $.89 \;\; (.04)$ & $.88 \;\; (.02)$ & $.88 \;\; (.01)$ & $.15 \;\; (.06)$ & $\num{1e07}$ \\
      \midrule
      \texttt{LLaMA-Guard-3-1B} & .88 & .95 & .49 & .65 & .03 & $\num{1e09}$\\ 
      \texttt{LLaMA-Guard-3-8B} & .89 & .95 & .52 & .68 & .03 &  $\num{8e09}$ \\
      \texttt{LLaMA-Guard-2-8B-FT} & -- & -- & -- & .85 & .10 & $\num{8e09}$ \\
      \bottomrule
  \end{tabular}
    \caption{In-distribution \our{} BeaverTails-30k test set performance,
    trained with various chat models.
    Note that \our{} achieves good performance (better AUPRC, but
    with higher FPR) than competing models at 100-1000x reduced guard
    parameter overhead.
    \tableinfo{}
    LLaMA-Guard-2/3 training set may overlap with the
    BeaverTails-30k test split, see~\citep{metallamaguard2}.
    LLaMA-Guard-2-8B-FT is a full fine-tune of LLaMA-Guard-2-8B
    on BeaverTails-30k, using an alternative train/test split designed
    to prevent overlap with the LLaMA-Guard-2 training set~\citep{metallamaguard2}.
    The \our{} models are trained on BeaverTails-30k.}%
    \label{tab:beavertails}
\end{table*}

\section{The \our System}
\label{section:methodology}

A guard model $\mathcal{G}$ for a generative chat model $\mathcal{C}$ categorizes each input and/or corresponding output of $\mathcal{C}$ according to a taxonomy of harmfulness categories.
The taxonomy could include coarse-grained categories, such as safe and unsafe, or could further distinguish between fine-grained categories, such as violence, hate, illegal 
activities, etc.

We now introduce \our. We assume a chat model $\mathcal{C}$ consisting of an embedding $\phi$, a feature map $\lmf$ and a linear language modelling head $\lmh$.
The embedding maps tokens to vectors, the feature map (a transformer variant;~\citealp{vaswani2017attention}) maps these vectors into representations and the language modelling head maps these representations into next-token logits. 
If $x$ represents a tokenized input sequence, then the next token logits are computed by $\lmh(f(\phi(x)))$.
We propose to build the guard model $\mathcal{G}$ using parameter-efficient fine-tuning methods applied to $\lmf$, and instantiate this idea with LoRA adapters, which add additional training parameters in the form of low-rank (i.e.~parameter-efficient) matrices (see~\cref{sec:related} for details). Other adaptation methods are possible~\citep{sung2022lst,he2021towards,lialin2023scaling,houlsby2019parameter}.

The same tokenizer and embedding is used for $\mathcal{C}$ and $\mathcal{G}$. 
However, $\mathcal{G}$ uses a different feature map $\gmf$ chosen as LoRA adapters attached to $\lmf$, and uses a separate output head $\gmh$ (linear, with bias), which maps features to a safe/unsafe logit and logits for harmfulness categories.
Tokenized content $x$ is therefore classified using $\gmh(\gmf(\phi(x)))$.
Deactivating the LoRA adapters and using the language modelling head gives the original chat model, while activating the LoRA adapters and using the guard model head gives the guard model. These \emph{generative} and \emph{guarding} paths, respectively, are depicted in Figure~\ref{fig:peft-guard-diagram}.
We do not merge the LoRA adapters into the chat
model weights after training, which allows for  dual use.

The dual path design of \our opens the door to methods based on adaptation instead of safety alignment fine-tuning.
Adaptation has an important advantage over safety alignment fine-tuning: the generative task is unaffected, 
so \our{} avoids any performance degradation on the generative task from safety fine-tuning 
(e.g., catastrophic forgetting;~\citealp{luo2023empirical}).

Most parameters, namely those in $\lmf$, are shared between the generative and guarding paths.
Therefore, the parameter overhead incurred by the guard model is only that of the LoRA adapters $\gmf$ and of the guard output head $\gmh$.
This is a tiny fraction of the number of parameters used by the chat system, often 3 orders of magnitude smaller, as shown in~\cref{tab:beavertails}.
We stress that deactivating the LoRA adapters and activating the language modelling head recovers exactly the original chat model, so no loss in chat performance is possible. 

The guard model is trained by supervised fine-tuning $\gmf$ and $\gmh$ on a
dataset labelled according to the chosen taxonomy.
Datasets are discussed in~\cref{sec:methods}.
During training, the parameters of the chat model $f$ remain frozen. Thereby, adapters of $\mathcal{G}$ are trained to leverage existing knowledge in $\mathcal{C}$.

\section{Methods}\label{sec:methods}

\subsection{Chat Models}
\label{sec:chatmodels}

We evaluate \our by training our guard adaptations with 3 small chat models covering
a range of sizes:
LLaMA-3.1-8B-Instruct~\citep{llama3.1modelcard},
LLaMA-3.2-1B-Instruct and LLaMA-3.2-3B-Instruct~\citep{llama3.2modelcard}.
We use the instruction tuned variants of each model to replicate their dual use as chat applications.
We use the PyTorch model implementations provided by the HuggingFace \texttt{transformers} library~\citep{wolf2019huggingface}.
We use the LoRA adapters provided in the HuggingFace PEFT module~\citep{peft2022}.

\subsection{Data}
The Beavertails dataset is used for training, the others for out of
distribution evaluation.
Datasets are accessed using the 
HuggingFace datasets module~\citep{lhoest2021datasets}.

\subsubsection{Beavertails-30k}\label{sec:beavertails}

The Beavertails-30k dataset~\citep{ji2024beavertails}
consists of prompt-response pairs, constructed using
the Alpaca-7B model~\citep{alpaca} to generate multiple unique responses
to 7774 unique prompts extracted from the 
HH Red-Team dataset~\citep{ganguli2022red}%
\footnote{\url{https://huggingface.co/datasets/PKU-Alignment/BeaverTails}}.
The resulting 30,207 prompt-response pairs are labelled by crowdworkers as
based on whether they belong (non-exclusively) to the 14 harm categories:
\begin{itemize*}
\item Hate Speech, Offensive Language
\item Discrimination, Stereotype, Injustice
\item Violence, Aiding and Abetting, Incitement
\item Financial Crime, Property Crime, Theft
\item Privacy Violation
\item Drug Abuse, Weapons, Banned Substance
\item Non-Violent Unethical Behavior
\item Sexually Explicit, Adult Content
\item Controversial Topics, Politics
\item Misinformation Re. ethics, laws and safety
\item Terrorism, Organized Crime
\item Self-Harm
\item Animal Abuse
\item Child Abuse.
\end{itemize*}

If a prompt-response pair belongs to any of these harm categories it is said to be
harmful. If it belongs to none of these harm categories it is said to be safe.
Descriptions of the harm categories are given in~\cref{sec:harm-categories}.
Within the dataset prompts are repeated (getting different model responses)
and sometimes prompt-response pairs are repeated (labelled differently 
by different crowd workers).

The dataset consists of train and test splits of sizes 27186 and 3021 respectively.
We further subdivide the train split into training and validation splits of
sizes 24672 and 2514 respectively.
We format examples using the template:
\texttt{user: \{prompt\}}
\texttt{<newline>}
\texttt{<newline>} 
\texttt{agent: \{response\}}.
Examples are right-padded for training and 
right truncated if necessary 
to fit within the model context window.

\subsubsection{ToxicChat}
The ToxicChat dataset~\citep{lin2023toxicchat} consists of 
$10,165$ prompt-response pairs from the 
Vicuna online demo~\citep{lin2023toxicchat,vicuna2023},
each annotated with a binary toxicity label corresponding to whether
the example is harmful/safe.
The examples are formatted in the same way as for BeaverTails-30k.

We use the January 2024 (0124) version available on 
HuggingFace.\footnote{\href{https://huggingface.co/datasets/lmsys/toxic-chat}{https://huggingface.co/datasets/lmsys/toxic-chat}}
The dataset is provided in a split of 5082 training examples and 5083 test examples.
We use the test set for evaluation.

\subsubsection{OpenAI Moderation}
The OpenAI Moderation Evaluation dataset~\citep{markov2023holistic}
consists of $1,680$ prompts (no model responses)
collected from publicly available sources,
labelled according to a taxonomy with $8$ harm categories.%
\footnote{\url{https://huggingface.co/datasets/mmathys/openai-moderation-api-evaluation}}
The dataset was used as an evaluation dataset by~\citet{markov2023holistic} to assess the performance of the OpenAI moderation API.
For evaluation we do not use the harm categories, and instead use an overall
harmfulness label. An example is harmful if it has a positive label in
any of the harm categories and safe otherwise.%
\footnote{The 8 categories determining harmful content are {sexual},
{hate}, {violence}, {harassment}, {self-harm},
{sexual/minors}, {hate/threatening} and {violence/graphic}.}
The prompts are formatted as \verb|user: {prompt}| before being passed to 
the guard model.

This dataset contains null labels.
We remove ambiguous examples where there are null labels in some categories
and no positive labels in any others.
These examples would have their overall harmfulness label determined by 
whether we view the null values as safe or unsafe.

\subsection{Training and Implementation}\label{sec:impl}
We train the guard models using the \our method on top of each
of the chat models specified earlier.
Training is performed on 8 NVIDIA A40s using data parallel with per-device batch size of 2, right padding and 2 gradient accumulation steps (resulting in batch size 32).
All computation is done in the PyTorch 16 bit brain float data type \texttt{bfloat16}.
For multi-GPU training with data parallel and gradient accumulation
we use the HuggingFace accelerate package~\citep{accelerate2022}.

For each of the models we use batch size of 32, LoRA rank $r=32$, scaling 
parameter $\alpha=64$ and LoRA dropout $0.05$.%
\footnote{We set $\alpha=2r$ following standard practice,
e.g., see~\citet{raschka2023lora}.}
LoRA adaptation is applied only to the query and key values of attention parameters in the chat models (no other layers or parameters
are adapted).

We initialise the guard model output heads using Xavier uniform initialisation~\citep{glorot2010understanding}.
In the notation of~\cref{sec:lora}, we initialise the LoRA parameters by setting $B$ to 0 and using Kaiming uniform initialisation~\citep{he2015delving} for $A$.
We train the model for 30 epochs
using AdamW~\citep{loshchilov2017decoupled} with learning
rate~\num{3e-4}. (We find that far fewer epochs are needed for
good validation performance.)
Each run is performed for 3 independent random seeds (which determine
the train/validation splits and the initialisation).

The loss function is the sum of loss functions for the safe/unsafe head
and the harmfulness category output heads, with each of these two terms
having equal weight 0.5:
$\ell = (\ell_{\text{unsafe}} + \ell_{\text{category}}) / 2$.
The loss for the safe/unsafe label is binary
cross-entropy while for the harmfulness categories we use a multi-label cross-entropy
loss with equal weights between categories.%
\footnote{Equivalently, with $n$ categories ($n=14$ training on BeaverTails)
there are $n + 1$ output heads each with an independent
binary cross-entropy loss.
The overall loss is a weighted sum,
the term corresponding to the safe/unsafe head with weight 0.5
and each of the category heads having weight $(2n)^{-1}$.}
The positive terms in each of the losses are further weighted
by the ratio of the number of
negative examples to that of positive examples in the training split. 

\subsection{Evaluation}
At the end of each epoch we perform a sweep across the entire train,
validation and test splits calculating various performance metrics 
with a classification threshold of~$0.5$.
We use the model checkpoint (end of epoch)
with the highest score for area under the precision recall curve (AUPRC)
on the validation set.
We report the median and min/max range over random seeds for each metric
on the test set.
We provide evaluations of LLaMA-Guard models~\citep{metallamaguard2,dubey2024llama3herdmodels}
on the datasets we consider.
We calculate these results by identifying the probability of the token for
\texttt{unsafe} after stripping preceding newline tokens from the response.

\section{Results}\label{sec:results}

\subsection{Beavertails-30k}
In~\cref{tab:beavertails} we have the test set results of \our{} trained on top of 
the three chat models considered, compared to baselines from the literature.
Note that performance is on par with LLaMA-Guard-2-8B-FT, which is a fine-tune
of LLaMA-Guard 2~\citep{metallamaguard2} (which has 8B parameters) on BeaverTails-30k.
Due to the dual-path design, the \our{} methods provide this strong performance while
incurring 100-1000x smaller guard parameter overhead when considered in conjunction
with the chat system being monitored.
Area under the precision-recall curve (AUPRC) exceeds that of the more recent
LLaMA-Guard 3 models~\citep{dubey2024llama3herdmodels}, but the LLaMA-Guard 3
models provide a lower false positive rate.
In~\cref{tab:category-auprc,tab:category-fpr} we
provide \our{} performance breakdown by harm category.

\begin{table*}
  \begin{tabular}{llll}
    \toprule
    & \our{}- & \our{}- & \our{}- \\
      Harm Category & \texttt{LLaMA-3.1-8B} & \texttt{LLaMA-3.2-3B} & \texttt{LLaMA-3.2-1B} \\
    \midrule
    Animal Abuse 
& $.71 \;\; (.09)$ 
& $.75 \;\; (.06)$ 
& $.70 \;\; (.10)$ 
\\
    Child Abuse 
& $.87 \;\; (.08)$ 
& $.87 \;\; (.04)$ 
& $.82 \;\; (.10)$ 
\\
    Controversial Topics, Politics 
& $.47 \;\; (.07)$ 
& $.52 \;\; (.03)$ 
& $.54 \;\; (.07)$ 
\\
    Discrimination, Stereotype, Injustice 
& $.81 \;\; (.02)$ 
& $.82 \;\; (.01)$ 
& $.81 \;\; (.02)$ 
\\
    Drug Abuse, Weapons, Banned Substance 
& $.77 \;\; (.03)$ 
& $.77 \;\; (.04)$ 
& $.77 \;\; (.03)$ 
\\
    Financial Crime, Property Crime, Theft 
& $.78 \;\; (.02)$ 
& $.79 \;\; (.02)$ 
& $.80 \;\; (.04)$ 
\\
    Hate Speech, Offensive Language 
& $.72 \;\; (.05)$ 
& $.76 \;\; (.01)$ 
& $.74 \;\; (.05)$ 
\\
    Misinformation Regarding Ethics, Laws, Safety 
& $.24 \;\; (.08)$ 
& $.20 \;\; (.10)$ 
& $.17 \;\; (.09)$ 
\\
    Non Violent Unethical Behavior 
& $.72 \;\; (.04)$ 
& $.72 \;\; (.02)$ 
& $.70 \;\; (.03)$ 
\\
    Privacy Violation 
& $.87 \;\; (.02)$ 
& $.87 \;\; (.03)$ 
& $.87 \;\; (.01)$ 
\\
    Self Harm 
& $.70 \;\; (.18)$ 
& $.68 \;\; (.11)$ 
& $.69 \;\; (.16)$ 
\\
    Sexually Explicit, Adult Content 
& $.68 \;\; (.04)$ 
& $.72 \;\; (.10)$ 
& $.68 \;\; (.03)$ 
\\
    Terrorism, Organized Crime 
& $.32 \;\; (.17)$ 
& $.27 \;\; (.02)$ 
& $.27 \;\; (.05)$ 
\\
    Violence, Aiding, Abetting, Incitement 
& $.85 \;\; (.02)$ 
& $.86 \;\; (.01)$ 
& $.84 \;\; (.02)$ 
\\
    \bottomrule
  \end{tabular}
  \caption{
Area under precision-recall curve by harm category for 
\our{} BeaverTails-30k test split, trained on top of various chat models.
Predictions made using the dedicated output head
per harm category in the guard model.
We report the median value across 3 random seeds with the range in parentheses.
Performance is generally good across categories but not matching the ability of the
model in predicting the overall safe/unsafe label of the examples.
This suggests that there is utility in having the dedicated safe/unsafe output head.%
}\label{tab:category-auprc}
\end{table*}

\begin{table*}
  \begin{tabular}{lrrr}
    \toprule
    & \our{}- & \our{}- & \our{}- \\
      Harm Category & \texttt{LLaMA-3.1-8B} & \texttt{LLaMA-3.2-3B} & \texttt{LLaMA-3.2-1B} \\
        \midrule
    Animal Abuse 
& $.02 \;\; (.02)$ 
& $.02 \;\; (.01)$ 
& $.01 \;\; (.01)$ 
\\
    Child Abuse 
& $.01 \;\; (.01)$ 
& $.02 \;\; (.01)$ 
& $.01 \;\; (.01)$ 
\\
    Controversial Topics, Politics 
& $.08 \;\; (.09)$ 
& $.08 \;\; (.07)$ 
& $.06 \;\; (.03)$ 
\\
    Discrimination, Stereotype, Injustice 
& $.07 \;\; (.08)$ 
& $.08 \;\; (.02)$ 
& $.06 \;\; (.02)$ 
\\
    Drug Abuse, Weapons, Banned Substance 
& $.05 \;\; (.07)$ 
& $.05 \;\; (.01)$ 
& $.04 \;\; (.03)$ 
\\
    Financial Crime, Property Crime, Theft 
& $.06 \;\; (.06)$ 
& $.08 \;\; (.04)$ 
& $.06 \;\; (.03)$ 
\\
    Hate Speech, Offensive Language 
& $.12 \;\; (.09)$ 
& $.09 \;\; (.05)$ 
& $.08 \;\; (.05)$ 
\\
    Misinformation Regarding Ethics, Laws, Safety 
& $.08 \;\; (.08)$ 
& $.07 \;\; (.08)$ 
& $.08 \;\; (.06)$ 
\\
    Non Violent Unethical Behavior 
& $.15 \;\; (.16)$ 
& $.13 \;\; (.05)$ 
& $.16 \;\; (.03)$ 
\\
    Privacy Violation 
& $.04 \;\; (.04)$ 
& $.04 \;\; (.02)$ 
& $.03 \;\; (.02)$ 
\\
    Self Harm 
& $.02 \;\; (.02)$ 
& $.05 \;\; (.02)$ 
& $.01 \;\; (.01)$ 
\\
    Sexually Explicit, Adult Content 
& $.02 \;\; (.05)$ 
& $.03 \;\; (.02)$ 
& $.03 \;\; (.01)$ 
\\
    Terrorism, Organized Crime 
& $.02 \;\; (.04)$ 
& $.07 \;\; (.06)$ 
& $.03 \;\; (.04)$ 
\\
    Violence, Aiding, Abetting, Incitement 
& $.12 \;\; (.05)$ 
& $.10 \;\; (.05)$ 
& $.10 \;\; (.02)$
\\
    \bottomrule
  \end{tabular}
  \caption{%
False positive rate by harm category for 
\our{} BeaverTails-30k test split, trained on top of various chat models.
Predictions made using the dedicated output head
per harm category in the guard model.
We report the median value across 3 random seeds with the range in parentheses.
False positive rate is low across the majority of categories.%
 }\label{tab:category-fpr}
\end{table*}

\subsection{OpenAI Moderation}
In~\cref{tab:openai} we report out of distribution evaluation of \our{}
on the OpenAI Moderation Evaluation dataset.
We see that the  AUPRC is close to that of the much (100-1000x) more costly LLaMA-Guard 3 
models, but with a considerably higher false positive rate. This is likely due to the low
bar for an example to be considered unsafe in BeaverTails~\citep{metallamaguard2}.
One might seek to mitigate this by training on a custom dataset.

\begin{table*}[h]
\centering
  \begin{tabular}{lllllll}
    \toprule
    Model & AUPRC$\uparrow$  & Precision$\uparrow$ & Recall$\uparrow$ & F1$\uparrow$ & FPR$\downarrow$ & Size$\downarrow$\\
     \midrule
\our{}-\texttt{LLaMA-3.2-1B} & $.85 \;\; (.02)$ & $.82 \;\; (.01)$ & $.54 \;\; (.12)$ & $.65 \;\; (.08)$ & $.52 \;\; (.16)$ & $\num{3e06}$ \\
\our{}-\texttt{LLaMA-3.2-3B} & $.86 \;\; (.02)$ & $.83 \;\; (.05)$ & $.51 \;\; (.22)$ & $.63 \;\; (.19)$ & $.47 \;\; (.29)$ & $\num{9e06}$ \\
\our{}-\texttt{LLaMA-3.1-8B} & $.86 \;\; (.03)$ & $.83 \;\; (.02)$ & $.56 \;\; (.20)$ & $.67 \;\; (.12)$ & $.51 \;\; (.29)$ & $\num{1e07}$ \\
\midrule
\texttt{LLaMA-Guard-3-1B} & .88 & .79 & .80 & .80 & .34 & $\num{1e09}$\\ 
\texttt{LLaMA-Guard-3-8B} & .94 & .88 & .79 & .83 & .17 & $\num{8e09}$\\ 
\bottomrule
\end{tabular}
\caption{Out-of-distribution evaluation of \our{} on OpenAI Moderation Evaluation dataset.
\tableinfo{}
AUPRC is close to that of LLaMA-Guard 3 at 100-1000x reduced guard overhead, but
false positive rate is considerably higher. This is likely due to the low
bar for an example to be considered unsafe in BeaverTails~\citep{metallamaguard2} 
and may be mitigated by training on an alternative dataset.}\label{tab:openai}
\end{table*}

\subsection{ToxicChat}
In~\cref{tab:toxic-chat} we provide out of distribution performance results 
of \our{} on the ToxicChat dataset, along with an evaluation of the LLaMA-Guard 3 models and 
the Open AI Moderation API~\citep{markov2023holistic} as baselines.
We see that the performance of all models is quite poor on this dataset.
We hypothesise that ToxicChat is a large distribution shift from BeaverTails-30k,
on which the \our{} models are trained.
We expect that the examples in the LLaMA-Guard training set have a similar
distribution to those in BeaverTails, since both datasets are originally
constructed from Anthropic's Red Team/Helpful-Harmless datasets.

The LLaMA-Guard 3 chat template indicates that the model should judge the safety of the final
turn in the conversation.
Inspecting ToxicChat examples, we found many with unsafe prompts but model refusals.
We performed the evaluation again moderating only the prompts in the examples, but saw
only a mild increase in performance.
ToxicChat also contains specific jailbreak attempts, but removing these also did not
improve performance.
    
\begin{table*}[h]
  \begin{tabular}{lllllll}
    \toprule
    Model & AUPRC$\uparrow$  & Precision$\uparrow$ & Recall$\uparrow$ & F1$\uparrow$ & FPR$\downarrow$ & Size$\downarrow$\\
     \midrule
\our{}-\texttt{LLaMA-3.2-1B} 
& $.30 \;\; (.03)$ 
& $.29 \;\; (.08)$ 
& $.38 \;\; (.10)$ 
& $.36 \;\; (.05)$ 
& $.07 \;\; (.04)$ 
& $\num{3e06}$ \\
\our{}-\texttt{LLaMA-3.2-3B} 
& $.40 \;\; (.03)$ 
& $.53 \;\; (.16)$ 
& $.40 \;\; (.22)$ 
& $.46 \;\; (.15)$ 
& $.03 \;\; (.02)$
& $\num{9e06}$ \\
\our{}-\texttt{LLaMA-3.1-8B} 
& $.36 \;\; (.07)$ 
& $.37 \;\; (.15)$ 
& $.41 \;\; (.16)$ 
& $.39 \;\; (.03)$ 
& $.05 \;\; (.06)$
& $\num{1e07}$ \\
\midrule
\texttt{LLaMA-Guard-3-1B} 
& .17 
& .15 
& .25 
& .18 
& .11
& $\num{1e09}$\\ 
\texttt{LLaMA-Guard-3-8B} 
& .32 
& .52 
& .21 
& .30 
& .02 
& $\num{8e09}$\\ 
\texttt{OpenAI Moderation API}  &$.63$ & $.55$  & $.70$ & $.61$ & --- \\
\bottomrule
\end{tabular}
\caption{Out-of-distribution evaluation of \our{} on ToxicChat dataset.
\tableinfo{}
The performance of all models is relatively poor on this dataset, with the OpenAI Moderation
API being the best.
We hypothesise that ToxicChat is a large distribution shift from the BeaverTails
dataset on which the \our{} models are trained. 
The performance of LLaMA-Guard 3 is also poor, possibly for a similar reason.
but their.
The OpenAI evaluations were performed on Jan 25 2024 using score threshold of 0.02,
results taken from \url{https://huggingface.co/lmsys/toxicchat-t5-large-v1.0}.%
}\label{tab:toxic-chat}
\end{table*}
\section{Related Work}
\label{sec:related}
\paragraph{Attacks.}
Jailbreak attacks have been shown to effectively generate harmful content \citep{rao2023tricking,kang2023exploiting}.
The overarching goal of a jailbreak attack
is to trick the model into ignoring or deprioritizing its safety mechanisms,
opening the door for the generation fo harmful content.

Simple approaches such as manual prompting are often 
effective~\citep{walkerspider2022,Mowshowitz2022,zswitten2022,Guzey2023,zeng2024johnny}.
Some example strategies include: 
instructing the model to ignore previous instructions
(aimed at circumventing safety instructions in the system prompt)~\citep{perez2022ignore,shen2023anything,schulhoff2023ignore};
asking the model to start the answer with ``\textit{Absolutely! Here's }'' to condition the generation process to 
follow a helpful direction~\citep{wei2024jailbroken}; 
using low-resource languages of alternative text modes such as ciphers, 
for which pre-training data exists but safety data may be lacking~\citep{yong2023low,barak2023another,yuan2023gpt,jiang2024artprompt}; 
inducing persona modulation or role-playing~\citep{shah2023scalable,yuan2023gpt};
using an LLM assistant to generate jailbreak prompts~\citep{WitchBOT2023,shah2023scalable};
or using iterative prompt refinement to evade safeguards~\citep{takemoto2024all,russinovich2024great}.

More complex approaches involve automatically generated prompts.
Automation can be achieved through LLM assistants which generate, modify or optimize prompts for 
jailbreaking~\citep{chao2023jailbreaking,mehrotra2023tree,shah2023scalable,yu2023gptfuzzer}.
Black-box optimization approaches rely solely on model outputs.
\citet{lapid2023open,liu2023autodan} use genetic algorithms, and 
\citet{mehrotra2023tree,takemoto2024all} use iterative refinement to optimize adversarial prompts.
White-box optimization approaches assume access to the target LLM and often rely on gradient information.
\citet{zou2023universal} use greedy coordinate gradient search to find a prompt suffix that causes LLMs to 
produce objectionable content.
\citet{zhu2023autodan} uses uses a dual-goal attack that is capable of jailbreaking as well as stealthiness,
thus avoiding perplexity filters which detect out of distribution text.
In between black-box and white-box there are also grey-box optimization approaches which use token probabilities \citep{andriushchenko2024jailbreaking,paulus2024advprompter}.

\paragraph{Defences.}
In addition to the development of safety alignment approaches~\citep{ouyang2022training,bai2022constitutional},
external defence mechanisms have been proposed to detect undesirable content---we will refer to these collectively as
guardrails~\citep{markov2023holistic,dong2024building}. 

Self-defence is an approach whereby an LLM is used to evaluate the safety of 
user-provided prompts or model-generated responses with an additional forward pass~\citep{helbling2023llm,wang2023selfguard,li2023rain}.
Self-reminders are pieces of text placed in system prompts which remind LLMs to answer according to safety guidelines 
before answering~\citep{xie2023defending}.
In-context learning can be used to strengthen defences without retraining or fine-tuning \citep{wei2023jailbreak,lin2023unlocking,zhou2024defending,varshney2023art}.
Perplexity-based filters are designed to detect jailbreaks built from out of distribution text~\citep{jain2023baseline,alon2023detecting}.
\citet{xie2024gradsafe} develep a method to detect unsafe prompts by scrutinizing the gradients of safety-critical parameters in LLMs.

A number of commercial solutions addressing safety exist, with varying degree of openness as to the methods employed, such as:
Nvidia's NeMo Guardrails~\citep{rebedea2023nemo}, OpenAI's Moderation API~\citep{openaiModeration}, GuardrailsAI~\citep{GuardrailsAI}, 
Perspective API~\citep{perspectiveAPI},
Protect AI~\citep{protectai2024llmguard}, 
Opaque~\citep{opaque}
and Enkrypt AI~\citep{enkryptai}.

The closest works to \our{} are LLaMA-Guard~\citep{inan2023llama} and Self-Guard~\citep{wang2023selfguard}.
The LLaMA-Guard is a fine-tune of LLaMA-2-7B model~\citep{touvron2023llama2} for harmful content detection.
LLaMA-Guard provides good performance, but at the cost of running a 7B guard model in addition to any chat system.
Self-Guard involved fine-tuning the chat model to review its responses and generate a tag which indicates the presence
of harmful content. The Self-Guard is parameter efficient, but modifying the chat model weights allows for the possibility of
forgetting~\citep{luo2023empirical}.

\paragraph{Parameter-Efficient Fine-Tuning.}
To address the increasing computational costs of fully fine-tuning LLMs,
\textit{parameter-efficient fine-tuning} methods have been proposed~\citep{he2021towards,lialin2023scaling}.
Selective fine-tuning selects a subset of the model parameters to be fine-tuned~\citep{donahue2014decaf,gheini2021cross}.
Prompt tuning prepends to the model input embeddings a trainable ``soft prompt'' tensor~\citep{lester2021power}.
Adapters add additional trainable parameters to existing layers while keeping the original parameters fixed~\citep{houlsby2019parameter}.
Low-rank adaptation (LoRA) involves adding a small number of trainable low-rank matrices to (some of) the model's weights,
without affecting the original model parameters~\citep{hu2021lora}. 
Ladder side-tuning disentangles the backwards pass of the original and new parameters for more efficient back-propagation~\citep{sung2022lst}.
\section{Conclusion}
\our provides guardrails for conversational systems at a vastly reduced parameter overhead when compared with
standard approaches (100-1000x fewer in our experiments).
Moreover, this reduction in memory requirements comes without loss of chat performance 
and with moderation performance competitive with or surpassing the state of the art on benchmark tasks.
These are due, respectively, to a dual-path design and the knowledge sharing in parameter-efficient fine-tuning.
We consider \our to be an important contribution to guardrail methods for resource-constrained 
settings such as on-device LLMs.
\paragraph{Potential Risks}
Distribution shift presents a risk to any guardrail system.
Harmful content at test-time which is significantly
different from that 
that which the model was trained on (e.g., an entirely
new category of harmful content) may bypass safety filters.
This risk can be mitigated by further work to improve out of distribution generalisation, for instance on building richer
datasets.
\section{Limitations}
\our{} requires access to the chat model weights, so is only applicable in these cases and cannot be applied to black-box systems.
In addition, we train \our{} with a fixed taxonomy for
harm categories (matching those in Beavertails-30k),
so adaptation to different taxonomies would requires retraining.
This is in contrast to LLaMA-Guard, which in principle can adapt to new taxonomies via in-context learning.
It is possible to train a guard model in the \our framework to have this adaptability.
We leave an evaluation of this to future work.
\section{Ethical Considerations}
The choice of taxonomy for harmful content
presents an important ethical consideration.
The perceived harm of certain content may vary across groups or societies,
so the taxonomy used must be customised both to the application and the audience.
We advise caution when deploying general-purpose guardrails across multiple cultural and demographic groups.

Our method may contribute to a wider adoption of content-moderated LLMs, in particular enabling on-device moderation in resource-constrained settings due to the reduction in memory overhead of the guard model.

We comply with licence conditions for all pre-trained models and datasets used in the work.
Where relevant, we comply with intended use for derivative work.
\bibliography{references}
\appendix
\section{Low Parameter Baselines}\label{sec:low-parameter-baselines}

In~\cref{tab:low-parameter-baselines} we provide results of methods which
use fewer additional guard parameters than \our{}.
These baselines are designed to indicate whether 
the additional guard model parameters in the LoRA adapters provide a benefit.
By comparison with~\cref{tab:beavertails} we find that this is indeed
the case. 
In these results, as with \our{}, we use the instruction tuned variants of each the models.

\paragraph{Self Defense}
The LLM Self Defense approach was suggested by~\citet{phute2024selfdefense}.
We incorporate the prompt-response pair to be classified 
into an instruction prompt with the format shown 
~\cref{fig:self-defense}
\begin{figure*}
\lstset{
  columns=flexible,
  basicstyle=\ttfamily,
}
\begin{center}
\begin{lstlisting}
Question: Does this describe harmful behaviour?
user: {prompt}
agent: {response}
Respond with 'yes' or 'no' and give an explanation.
Answer: 
\end{lstlisting}
\end{center}
\caption{Prompt format for Self Defense baseline.}%
\label{fig:self-defense}
\end{figure*}
This text is them fed to the LLM and the generated outputs are parsed into a binary prediction.
Before processing the generated text we strip whitespace and convert to lower case.
If the processed output matches the regex
\verb|^yes[^a-zA-Z]| then the example
gets the binary prediction unsafe,
if it matches \verb|^no[^a-zA-Z]| then the example
gets the binary prediction safe.
If the example matches neither pattern
then we arbitrarily 
consider prediction to be unsafe.
The percentage of such examples is $45\%$ for LLaMA-3.2-1B,
$0\%$ for LLaMA-3.2-3B and $19\%$ for LLaMA-3.1-8B.
The results are not dramatically improved if we consider these examples as safe.

\paragraph{Output Head Tuning}
We perform an ablation study on \our{} in which we train only the
output head (without any LoRA adaptation) for the three chat models considered. 
The training configuration is the same as described in~\cref{sec:methods},
only without the LoRA adatpers, gradient accumulation
or data parallel 
(each run fits easily on one GPU).

\begin{table*}
  \begin{tabular}{lllllll}
    \toprule
    Model & AUPRC$\uparrow$  & Precision$\uparrow$ & Recall$\uparrow$ & F1$\uparrow$ & FPR$\downarrow$ & Size$\downarrow$\\
     \midrule
    Self-Defense-LLaMA-3.2-1B &  -- &
    $.55$ & $.72$ & $.62$ & $.78$  & $\num{0}$ \\
    Self-Defense-LLaMA-3.2-3B & -- &
    $.57$ & $.99$ & $.72$ & $.99$ & $\num{0}$ \\
    Self-Defense-LLaMA-3.1-8B & -- &
    $.59$ & $.77$ & $.67$ & $.71$ & $\num{0}$ \\
     Output-Head-LLaMA-3.2-1B & $.91 \;\; (.00)$ & $.76 \;\; (.00)$ & $.91 \;\; (.02)$ & $.83 \;\; (.01)$ & $.38 \;\; (.01)$ & $\num{2.87e04}$ \\
     Output-Head-LLaMA-3.2-3B  & $.92 \;\; (.01)$ & $.79 \;\; (.00)$ & $.91 \;\; (.01)$ & $.84 \;\; (.00)$ & $.33 \;\; (.01)$ & $\num{4.30e04}$ \\
     Output-Head-LLaMA-3.1-8B & $.93 \;\; (.01)$ & $.80 \;\; (.00)$ & $.91 \;\; (.01)$ & $.85 \;\; (.00)$ & $.30 \;\; (.01)$ & $\num{5.74e04}$ \\
      \bottomrule
  \end{tabular}
    \caption{Low parameter baseline results on BeaverTails-30k.
    By comparison with~\cref{tab:beavertails}, we see that the LoRA adapters provide
    a benefit for guard model performance.
    The Output-Head models are trained on 
    BeaverTails-30k but the Self-Defense ones
    are not (they are simply an in-context prompt asking if an example is unsafe.)
    See~\cref{sec:low-parameter-baselines} for details.
    \tableinfo{}%
    }\label{tab:low-parameter-baselines}
\end{table*}
\section{BeaverTails Harm Categories}\label{sec:harm-categories}
\begin{center}
    \emph{This section is copied verbatim from~\citep[Appendix B]{ji2024beavertails}, displayed here for readers' convenience.}
\end{center}

\textbf{Hate Speech, Offensive Language:} This category encompasses any form of communication or content that expresses intense hostility or violence towards an individual or group based on aspects like race, religion, ethnicity, nationality, gender, sexual orientation, or any other defining characteristics. This also includes language that is considered vulgar, crude, or offensive.

\textbf{Discrimination, Stereotype, Injustice:} Discrimination refers to treating people unfairly or unequally based on their personal characteristics or affiliations such as race, age, gender, religion, etc. Stereotyping involves holding simplified and generalized beliefs about particular groups, which often leads to prejudices and discrimination. Injustice pertains to unfair treatment or violation of the rights of individuals or groups.

\textbf{Violence, Aiding and Abetting, Incitement:} This involves any form of physical harm, threat, or violent behavior towards individuals or groups. Aiding and abetting refers to the act of helping, supporting, or encouraging such violent behaviors or illegal activities. Incitement pertains to the act of provoking or stirring up harmful, violent, or illegal actions.

\textbf{Financial Crime, Property Crime, Theft:} This includes all forms of illegal activities that lead to financial loss such as fraud, embezzlement, money laundering, bribery, and other forms of corruption. Property crime and theft involve the act of taking someone's property without their consent.

\textbf{Privacy Violation:} This refers to actions that invade an individual's privacy, such as unauthorized access, use, disclosure, or collection of personal data. It also includes stalking, doxing, or any other form of intrusion into someone's personal life without their consent.

\textbf{Drug Abuse, Weapons, Banned Substance:} This category includes the misuse or abuse of illegal drugs, prescription drugs, or any other banned substances. It also involves the unauthorized or illegal possession, use, or trade of weapons.

\textbf{Non-Violent Unethical Behavior:} This refers to any behavior that is morally or ethically wrong but does not involve violence. Examples can include dishonest behaviors like lying, cheating, deceiving, spreading rumors, betraying trust, or any other form of unethical conduct.

\textbf{Sexually Explicit, Adult Content:} This category includes any content that depicts explicit sexual activities or adult themes that are considered inappropriate or offensive. This can include pornography, explicit sexual text or messages, and any other form of sexually explicit material.

\textbf{Controversial Topics, Politics:} This refers to discussions or statements related to controversial topics, political ideologies, or sensitive issues that could lead to disagreement or conflict. These topics can be divisive and could potentially cause harm if not handled respectfully and thoughtfully.

\textbf{Misinformation Re. ethics, laws, and safety:} This category involves the spreading of incorrect or misleading information related to ethical issues, laws, regulations, or any matter related to safety. Misinformation can cause harm by influencing people's beliefs or actions negatively.

\textbf{Terrorism, Organized Crime:} This pertains to any form of content or action related to terrorism or organized crime, including endorsing or promoting terrorist activities, participating in organized criminal activities, or spreading propaganda for such groups.

\textbf{Self-Harm:} This includes any form of self-inflicted harm, whether physical or emotional, or any content that promotes, glorifies, or encourages self-harm. This can also include discussions of suicidal thoughts or plans.

\textbf{Animal Abuse:} This involves any form of cruelty or harm inflicted on animals, including physical abuse, neglect, or any behavior that causes distress or harm to an animal. It also includes content that promotes such behavior.

\textbf{Child Abuse:} This encompasses any form of physical, emotional, or sexual abuse directed toward children. It can also include neglect, exploitation, or any behavior that harms a child or violates their rights. Content that promotes or glorifies such behavior also falls under this category.

\end{document}